\documentclass[10pt,twocolumn,letterpaper]{article}

\usepackage{iccv}
\usepackage{times}
\usepackage{epsfig}
\usepackage{graphicx}
\usepackage{booktabs}
\usepackage{microtype}
\usepackage{subcaption}
\usepackage[dvipsnames]{xcolor}

\usepackage{amsmath}
\usepackage{amssymb}
\usepackage{bbold}
\usepackage{bm}
\usepackage{dsfont}

\usepackage{caption}
\usepackage{algorithm}
\usepackage{algpseudocode}
\usepackage{amsmath}

\makeatletter
\algnewcommand{\LineComment}[1]{\Statex \hskip\ALG@thistlm \(\triangleright\) #1}
\makeatother

\usepackage{booktabs}
\usepackage{tabularx}
\usepackage{multirow}

\usepackage{units}

\usepackage{pifont}
\newcommand{\cmark}{\ding{51}}%
\newcommand{\xmark}{\ding{55}}%

\DeclareMathOperator{\mixematch}{MixEMatch}
\DeclareMathOperator{\betadist}{Beta}

\iccvfinalcopy 

\def\iccvPaperID{12326} 

\ificcvfinal\fi

\newcommand{\COLOR}{black}

\begin{document}

\title{ Manifold DivideMix: A Semi-Supervised Contrastive Learning Framework for Severe Label Noise}


\author{
Fahimeh Fooladgar\\
University of British Columbia\\
\and
Minh Nguyen Nhat To\\
University of British Columbia\\
\and
Parvin Mousavi\\
Queen's University\\
\and
Purang Abolmaesumi\\
University of British Columbia\\
}

\maketitle

\ificcvfinal\thispagestyle{empty}\fi

\begin{abstract}
Deep neural networks have proven to be highly effective when large amounts of data with clean labels are available. However, their performance degrades when training data contains noisy labels, leading to poor generalization on the test set. Real-world datasets contain noisy label samples that either have similar visual semantics to other classes (in-distribution) or have no semantic relevance to any class (out-of-distribution) in the dataset. Most state-of-the-art methods leverage ID labeled noisy samples as unlabeled data for semi-supervised learning, but OOD labeled noisy samples cannot be used in this way because they do not belong to any class within the dataset. Hence, in this paper, we propose incorporating the information from all the training data by leveraging the benefits of self-supervised training. Our method aims to extract a meaningful and generalizable embedding space for each sample regardless of its label. Then, we employ a simple yet effective K-nearest neighbor method to remove portions of out-of-distribution samples. By discarding these samples, we propose an iterative ``Manifold DivideMix" algorithm to find clean and noisy samples, and train our model in a semi-supervised way. In addition, we propose ``MixEMatch", a new algorithm for the semi-supervised step that involves mixup augmentation at the input and final hidden representations of the model. This will extract better representations by interpolating both in the input and manifold spaces. 
Extensive experiments on multiple synthetic-noise image benchmarks and real-world web-crawled datasets demonstrate the effectiveness of our proposed framework. Code is available at  https://github.com/Fahim-F/ManifoldDivideMix.
\end{abstract}
\section{Introduction}
\begin{figure*}[t]
  \centering
  \includegraphics[width=1\textwidth]{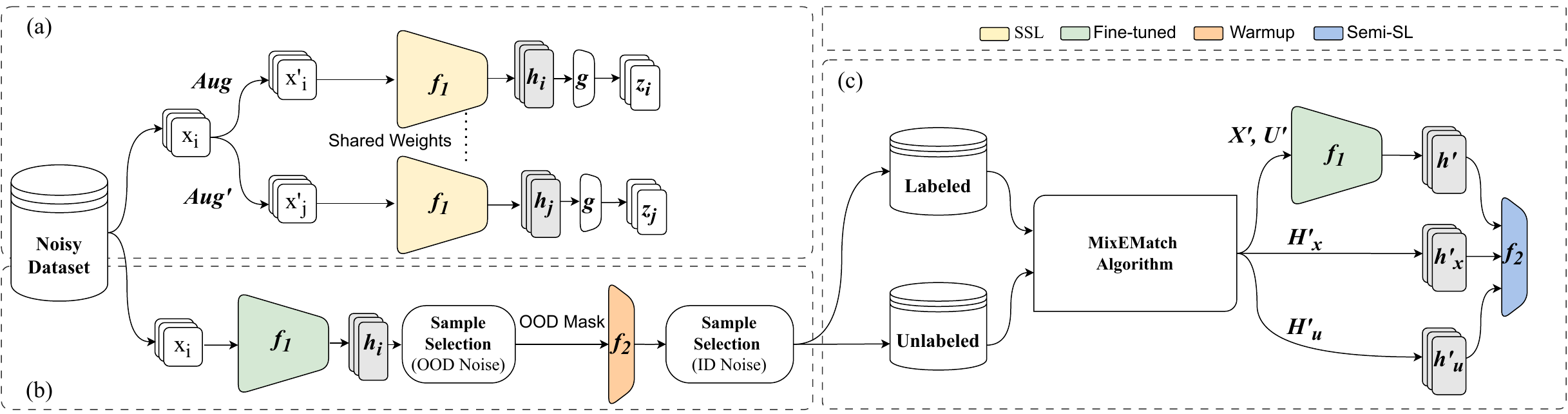}
\caption{An overview of the proposed method. (a) Step (1): Learn an embedding space of each image regardless of its label in the self-supervised way using contrastive loss  (\ref{eq:loss_self}) \cite{khosla2020supervised}. (b) Once, the backbone model trained,  in Step (2),  Select OOD samples based on the distances in the embedding space by using KNN and create OOD mask to remove OOD from supervised learning. Then, train a linear classifier ($f_2$) on top of the trained backbone ($f_1$) and fine-tune it for a few epochs. (c) Step (3): Select ID noisy samples based on the distribution of classification loss and find clean/noisy samples to form labeled/unlabeled sets. Next, continue training classifier ($f_2$) and fine-tuning backbone model ($f_1$) in the Semi-supervised way by using MixEMatch algorithm (\ref{alg:mixematch}). Step 2 and 3 repeated until convergence. The color legend shows how each component in each step has been trained.}
\label{fig:overall}
\end{figure*}
The empirical observation of deep neural networks (DNNs) applied to computer vision problems reveals that they produce superior performance when trained with a significant amount of clean, annotated data~\cite{hestness2017deep}. However, their primary weakness is also the vast number of clean labeled examples needed for training. Collecting and manually annotating such data could be complex, time-consuming, and costly, particularly in certain fields. It may also require domain expertise, such as in medical imaging. Meanwhile, open-source online data that can be automatically annotated using search engine queries and user tags is the backbone of most large-scale data collection methods~\cite{liu2021exploiting,yao2017exploiting}. However, this annotation approach will certainly result in label noise. Therefore, it is challenging to train DNNs using such data, as they can effectively memorize noisy random labels during training~\cite{2017_ICML_Memorization,2020_NeurIPS_EarlyReg}.

Several methods have been developed to deal with label noise in automatically annotated datasets, such as semi-supervised learning~\cite{2019_NeurIPS_MixMatch,sohn2020fixmatch}, self-supervised learning~\cite{2020_ICML_SimCLR}, and robust training~\cite{2020_ICLR_DivideMix}.

These methods can be classified into two primary categories. The first category assumes that the true labels of noisy samples are included in the label set (i.e. in-distribution labeled noisy samples (ID samples). The community has invested a lot of effort in designing robust methods to train DNNs in the presence of ID label noise~\cite{2020_ICLR_DivideMix,2018_NeurIPS_CoTeaching,2020_NeurIPS_EarlyReg}. The second category arises from the observation that techniques that are robust to ID noise tend to perform poorly when applied in more realistic settings (with real-world label noise). In fact, the authors of~\cite{2022_WACV_DSOS} suggest that most of the label noise in web-crawled datasets is out-of-distribution (OOD) label noise, which means that the true labels for noisy samples cannot be inferred from the distribution (we call them OOD samples). To evaluate this, they randomly collected three small but representative sample sets from the WebVision 1.0 dataset~\cite{2017_arXiv_WebVision} to determine the typical level of noise present in web-crawled, automatically annotated datasets. They reported that approximately 70\% of the data is properly labeled, 5\% have in-distribution noisy labels, and 25\% have out-of-distribution noisy labels. Hence, they argue that the primary barrier to removing label noise in web-crawled and any real-world dataset is related to the assumption that most label-noise-robust techniques only consider the ID noise scenario. 

Given the observations above, we need approaches to first separate potentially clean, ID noise and OOD noise instances, then decide how to incorporate those noisy data in the training. However, while ID noisy samples may be directly fed to the network as unlabeled data using common semi-supervised learning (Semi-SL) algorithms, OOD label noise samples cannot be assigned to any category or used to train the network.  Therefore, a simple solution is to remove them, once detected as OOD samples.  Albert et al.~\cite{2022_WACV_DSOS} propose a method to detect OOD labeled noisy samples and enforce their prediction to converge to a uniform distribution, rather than deleting the samples. 
Leveraging the OOD samples during unsupervised training improves the generality as these samples still include useful information for learning low-level features. As a result, they may be used to improve the representations that are acquired~\cite{2021_CVPR_JoSRC}. 
Although having both in- and out-of-distribution labeled noisy data presents challenges for supervised training of a model, these can be mitigated, and in some cases the data can even be advantageous, in a framework for self-supervised training. 

In this paper, we propose  ``Manifold DivideMix", which addresses learning with ID and OOD label noise in a semi-supervised way. Based on the success of Self-Supervised Learning (SSL) and different from most existing methods, we consider one model instead of two or ensemble model, and train it using unsupervised contrastive learning on the noisy dataset to learn meaningful representations of data without explicit labels.  Next, we add a linear classifier on top of the SSL model and incorporate two sample selection methods to detect OOD and ID samples gradually during training. As a results, we iteratively remove OOD samples and leverage the clean and ID samples in the Semi-SL training (see Figure 1). 
 In summary, our contributions are as follows.
\begin{itemize}
	\item 	
        Based on the assumption that measuring distances between images is more meaningful in the embedding space, we take the average distance between each embedding space and each of its K- Nearest Neighbours (KNN) as the OOD score to detect OOD labeled noisy samples;
	\item
         To learn better representation and improve the generalisation performance, we propose MixEMatch, an algorithm to apply mixup augmentation both on the input and representation spaces of the Semi-SL step. These augmentations boost the overall quality of pseudo-labels and, as a result, substantially improve the effectiveness of subsequent Semi-SL training;
	\item 
        We demonstrate experimentally that unsupervised feature learning reduces the effect of overfitting to label noise and significantly outperforms baseline approaches by a large margin, particularly when applied to datasets with high label noise.
\end{itemize}

\section{Related Work}

Song et al.~\cite{song2022learning} categorize the current research trends to address noisy label problem through: 1)  Robust regularisation~\cite{2018_ICLR_mixup,verma2019manifold}, which involves explicitly or implicitly forcing a DNN to overfit less to falsely labeled examples; 2) Robust loss functions~\cite{2019_ICCV_SL,zhou2021asymmetric} that can handle label noise; 3) Robust architectures~\cite{2017_ICLR_Smodel,yao2018deep} that have added noise adaptation layers to a DNN to learn a label transition process or a dedicated architecture to reliably support more diverse types of label noise; 4) Loss correction~\cite{patrini2017making} and loss reweighting~\cite{liu2015classification}, which modify the loss value based on the certainty of a particular loss or label; and 5) Sample selection, which extracts correctly labeled instances from noisy training data using multiple networks or iterative learning~\cite{2018_NeurIPS_CoTeaching,yu2019does}. {
Recently, based on the assumption that all noise in the data are either ID or a mix of ID and OOD label noise, there are two categories of research that address the following questions: 1) how can we detect noisy samples?, and 2) how do we leverage them during training once they are detected? 

\subsection{ID Sample Selection Methods}
 Many noise-robust methods assume that all the noise is ID. On the presumption that clean labels represent the majority in a noisy label dataset, deep networks initially remember training data with clean labels before moving on to data with noisy labels or complex samples. Hence, low-loss instances might be considered clean (highly probable occurrences). This approach has been successful in identifying potentially clean instances in a wide range of situations. To obtain the clean/noisy separation using unsupervised methods, a mixture model is fitted to the loss distribution of training samples. In DivideMix~\cite{2020_ICLR_DivideMix}, a Gaussian Mixture Mode (GMM) is used to detect high-loss samples. This concept is further developed in PropMix~\cite{2021_arXiv_propmix} by applying another GMM on the noisy samples to separate simple and hard noisy samples. Nevertheless, clean samples from the hard classes may generate high-loss values, hence the selection procedure is often skewed towards easy classes. This is especially noticeable at the beginning of training, and it might cause class-disparity among the chosen clean samples. Karim et al.~\cite{karim2022unicon} propose an efficient and scalable technique based on Jensen-Shannon divergence for separating samples, and choose an equal number of high-quality examples from each class to solve the problem of the over-selection of easy and complex samples. In~\cite{2018_NIPS_GoldLoss}, the trusted small clean set was considered to estimate the uncertainty of prediction and detect noisy labeled samples. 

\subsection{ID-OOD Sample Selection Methods}
 A new category of noise-resistant algorithms has recently emerged to deal with noisy datasets that include both in- and out-of-distribution label noise. In~\cite{2021_CVPR_JoSRC}, the authors utilized the Jensen-Shannon divergence between a predicted label and the original label to identify ID labeled noisy samples, and then samples with poor agreement across different views were considered OOD. 
 Albert et al.~\cite{albert2022embedding} proposed a method called SNCF that identifies clean, ID, and OOD samples by applying outlier sensitive clustering at the class-level. In~\cite{2021_ICCV_RRL}, the authors computed a pseudo-label by aggregating information from the top $K$ neighbors, and then selected a subset of training samples with reliable pseudo-labels as clean samples, assuming that images with several neighbors from the same class were less likely to be noisy. Meanwhile, in~\cite{2022_WACV_DSOS}, the authors proposed a method called Dynamic Softening for Out-of-distribution Samples (DSOS) to distinguish between ID and OOD samples by computing the collision entropy of the interpolation between the original label and network prediction. Another method, EvidentialMix~\cite{2020_WACV_EDM}, improved upon loss values clustering by using evidential loss~\cite{2018_NeurIPS_evidentialloss} to indicate the presence of two separate noisy modes, one for OOD and one for ID.

\subsection{Dealing with Label Noise}
Various methods have been developed to deal with noisy data once they are detected. The simplest way is to discard the potential either ID noise samples~\cite{2019_ICML_DiscardingBad,2018_NeurIPS_CoTeaching,yu2019does,2018_ICML_MentorNet} or OOD noise samples ~\cite{2020_WACV_EDM,2022_WACV_DSOS} during training. This leads to sub-optimal results as those samples still have helpful information to enhance generalization of the model. Most other methods handle the noisy samples as unlabeled data and continue training using common semi-supervised learning algorithms~\cite{2019_NeurIPS_MixMatch,sohn2020fixmatch,2020_ICLR_DivideMix}.
 In DivideMix~\cite{2020_ICLR_DivideMix} and UNICON~\cite{karim2022unicon}, two models were trained with a semi-supervised consistency regularization algorithm to incorporate noisy samples as unlabled samples and corrected their labels. This concept was further developed in PropMix~\cite{2021_arXiv_propmix}, where self-supervised initialization was employed and only the simple noisy samples were corrected, while the hardest ones were discarded. ScanMix~\cite{2021_arXiv_Scanmix} was an advancement on DivideMix that used a semi-supervised contrastive method to fix the label and semantic clustering in a self-supervised feature space.  The method in~\cite{2021_ICCV_RRL} encouraged clean samples to be similar to their class prototypes by training on a weakly supervised prototype. In order to overcome the challenges of label noise, the authors of~\cite{2021_CVPR_MOIT} presented a Multi-Objective Interpolation Training (MOIT) framework, where supervised contrastive learning and semi-supervised learning complemented one another. 
 \color{\COLOR}
The authors of ~\cite{yu2023prevent} propose Gradient Switching Strategy (GSS) to prevent the damage caused by mislabeled samples during training. They aim to eliminate the impact of misleading gradient directions of noisy labeled samples by assigning a random gradient direction for those samples.}

\section{Proposed Method}
Our proposed Manifold DivideMix algorithm (Figure~\ref{fig:overall}) begins with a self-supervised phase~\cite{khosla2020supervised,2020_ICML_SimCLR}. We next use filtering procedures in a semi-supervised training setup to first separate and eliminate the simple out-of-distribution samples. Afterwards, during the semi-supervised training phase, the remaining samples are determined iteratively and proportionately as clean, in- and out-of-distribution label noise samples. The following subsections provide a detailed explanation of each step. 

\subsection{Background} 
Let $\mathcal{D} = \left\{ \left(x_{i},y_{i}\right)\right\} _{i=1}^{N}$ represent the training set, where $\mathbf{x}_i \in \mathcal{X} $ is the $i^{th}$ sample, and $y_i $ denotes the given label associated with one of the $C$ classes, and $N$ is the total number of training samples. Specifically, we address the situation when the dataset contains $\mathcal{D}_{c}=\left\{ \left(x_{i},y_{i}\right)\right\} _{i=1}^{N_{c}}$, a properly labeled set, $\mathcal{D}_{in}=\left\{ \left(x_{i},y_{i}\right)\right\}_{i=1}^{N_{in}}$ as an incorrectly labeled in-distribution noisy set, and $\mathcal{D}_{out}=\left\{ \left(x_{i},y_{i}\right)\right\}_{i=1}^{N_{out}}$ as an out-of-distribution noisy set. Here, $N=N_{c}+N_{in}+N_{out}$ represents the total number of samples available. We assume that the sample distribution among $\mathcal{D}_{c}$, $\mathcal{D}_{in}$ and $\mathcal{D}_{out}$ is unknown. 
As a result of noise, the true label, $\hat{\mathbf{y}}_i$, may differ from the given label $\mathbf{y}_i$. We design the DNN model with an encoder base model, $f_1(.;\theta)$, a classification layer, $f_2(.;\phi)$, and a projection head, $g(.; \psi)$, with parameters $\theta$, $\phi$, and $\psi$, respectively.  We present an algorithm effective for training such a model on the corrupted annotated dataset $\mathcal{X}$ without overfitting to the noise with reliable classification of samples within the class distribution.


\subsection{Self-supervised Learning}
Initially, regardless of the noisy labels, the model is trained in a self-supervised way to learn representations from images. The goal is to extract rich feature representations not only to mitigate the effect of ID noise but also to capture clusters of similar images to identify OOD samples in the early stages. To achieve this, our algorithm learns representations by maximizing the agreement between two distinct augmented versions of the same image using a contrastive loss in the latent space. This approach is inspired by recent contrastive learning models~\cite{2020_ICML_SimCLR, khosla2020supervised}.

To train the model, two random data augmentations are applied to each training image $x_i$ to provide two related views of the same example. These views are considered as a positive pair in the contrastive learning algorithm and are denoted as ${\bm x^\prime}_i$ and ${\bm x^\prime}_j$. A neural network encoder, $f_1(\cdot;\theta)$, and a small neural network projection head, $g(\cdot;\psi)$, are constructed to map $\boldsymbol{x}$ to the representation space $\bm h_i = f_1({\bm x^\prime}_i;\theta) \in \mathbb{R}^{d}$, and $h_i$ to the $\bm z_i$ space, where the contrastive loss is applied (Figure~\ref{fig:overall}-(b)). The projection head is removed at inference time. Both augmented samples feed separately into the same encoder, resulting in a pair of representation vectors ($\bm h_i, \bm h_j$) and projection heads ($\bm z_i, \bm z_j$).
The loss function for each pair of examples $(i, j)$ is defined as self-supervised contrastive learning (e.g., \cite{khosla2020supervised}), as follows:

\begin{equation}
\label{eq:loss_self}
    \ell_{i,j} = -\log \frac{\exp(\mathrm{sim}(\bm z_i, \bm z_j)/\tau)}{\sum_{k=1}^{2M} \mathds{1}_{k \neq i}\exp(\mathrm{sim}(\bm z_i, \bm z_k)/\tau)},
\end{equation}
where $\mathrm{sim}(\bm z_i, \bm z_j)$ denotes the cosine similarity between $\bm z_i$ and $\bm z_j$,  $\mathds{1}_{k \neq i} \in \{ 0,  1\}$ is an indicator function evaluating $1$ if $k \neq i$, and $\tau$ denotes a temperature parameter. The final loss, $ \mathcal{L}^{self}
  =\sum_{i\in I}\mathcal{L}_i^{self}$, is computed across all augmented pairs in a mini-batch $I$, where $\mathcal{L}_i^{self}$ defined in Eq. (\ref{eq:loss_self}).

\subsection{Sample Selection}
Following self-supervised training, we first detect potential OOD samples based on the trained embedding space and remove them from the training set (Figure~\ref{fig:overall}-(b)). Additionally, we add a linear classifier, $f_2(.;\phi)$, on top of the learned representations for classification. We train the classifier for a few epochs on the whole training set with a symmetric cross-entropy loss~\cite{2019_ICCV_SL} to warm it up and make it robust to a high level of label noise, even at the beginning of training. Next, we divide the remaining samples into clean and noisy in-distribution sets to start the Semi-SL algorithm. We iteratively apply filtering to identify noisy samples (ID and OOD) during training, which is explained in the following subsections.

\textbf{OOD Label Noise Detection:} Rather than defining a new metric based on the predictions of the model and given noisy labels~\cite{2022_WACV_DSOS}, Generative Adversarial Networks~\cite{zhang2021understanding}, or autoencoder architecture~\cite{yang2021generalized}, we utilize a simple algorithm to detect OOD  samples by leveraging the $K$ nearest neighbours  of any example $x$ and their distance to $x$ as a primitive estimate of the local density around $x$~\cite{kuan2022back}. To determine the degree of OOD in a given image $x$, we calculate the average distance between $x$ and each of its $K$ nearest neighbours from the training data, which has demonstrated acceptable OOD detection results. 

\textbf{ID Label Noise Detection:} 
In the presence of ID label noise and prior to overfitting, we may anticipate that the properly classified samples are those with clean labels, while the remaining samples have noisy labels or are complicated ones. 
Let $p \left ( x_i=\text{clean} | \ell^{sup}_i , \gamma \right )$ be a function that estimates the probability that $(\mathbf{x}_i,\mathbf{y}_i)$ is a clean label sample based on $\ell^{sup}_i = -\mathbf{y}_i\log p_{\phi}(y_i|f_{\theta}(\mathbf{x}_i))$ which is the simple cross-entropy loss function.  The probability $p \left ( x_i=\text{clean} | \ell^{sup}_i, \gamma \right )$ can be estimated using a bimodal Gaussian mixture model (GMM)~\cite{2020_ICLR_DivideMix}. $\gamma$  represents the GMM parameters, where the component with the larger mean value is considered to be the noisy component, while the component with the lower mean value is considered to be the clean component. As a result, clean ($\mathcal{X} \subseteq \mathcal{D}$) and possibly ID labeled noisy samples ($\mathcal{U} \subseteq \mathcal{D}$) are formed as follows:
\begin{equation}
\begin{split}
    \mathcal{X}  = \left \{ (\mathbf{x}_i,\mathbf{y}_i, w_i) : w_i=p \left ( x_i=\text{clean} | \ell^{sup} , \gamma \right ) \ge \tau_2 \right \}, \\
    \mathcal{U}  = \left \{ (\mathbf{x}_i,\mathbf{y}_i, w_i) : w_i=
p \left ( x_i=\text{clean} | \ell^{sup} , \gamma \right ) < \tau_2 \right \},
\end{split}
\label{eq:clean_noisy_sets}
\end{equation}
where $\tau_2$ denotes a clustering threshold.

\begin{algorithm*}
    \caption{MixEMatch}
    
    \label{alg:mixematch}
    
    \begin{algorithmic}[1]
        \Require $\mathcal{X} = \big((x_b, y_b, w_b); b \in (1, \ldots, B)\big)$ \Comment{batch of labeled examples, their labels and probability of being clean sample}
        \Require $\mathcal{U} = \big(u_b, w_b; b \in (1, \ldots, B)\big)$ \Comment{batch of unlabeled examples and probability of being clean sample}
        \Require $T$, $\alpha$ \Comment{Sharpening temperature, Parameter of $\betadist$ distribution for Mixup}
        
        

        \Statex
        
        \For{$b \gets 1$ \textbf{to} $B$}
            \State $x^w_{b,1}, {x}^w_{b,2} \gets \text{Aug}^W(x_b, 2)$ \Comment{apply two weak  augmentations on $x_b$}
            \State $x^s_{b,1}, {x}^s_{b,2} \gets \text{Aug}^S(x_b, 2)$ \Comment{apply two strong  augmentations on $x_b$}
            \State $u^w_{b,1}, {u}^w_{b,2} \gets \text{Aug}^W(u_b, 2)$ \Comment{apply two weak  augmentations on $u_b$}
            \State $u^s_{b,1}, {u}^s_{b,2} \gets \text{Aug}^S(u_b, 2)$ \Comment{apply two strong  augmentations on $u_b$}

            \Statex
            
            \State ${p}_{b} \gets \frac{1}{2}\big(\mathrm{p}({x}^w_{b,1};\theta, \phi) + \mathrm{p}({x}^w_{b,2};\theta, \phi)\big)$ \Comment{average the predictions across augmentations of $x_b$}
            \State $\bar{y}_b \gets w_b y_b+(1-w_b){p}_{b}$ \Comment{refine ground-truth label guided by the clean probability produced by the GMM}
            \State ${y}_b \gets \text{Sharpen}(\bar{y}_b,T)$ \Comment{apply temperature sharpening to the refined label}

            \Statex
            
            \State $\bar{q}_{b} \gets \frac{1}{2}\big(\mathrm{p}({u}^w_{b,1};\theta, \phi) + \mathrm{p}({u}^w_{b,2};\theta, \phi)\big)$ \Comment{guessing the label for unlabeled sample by averaging the predictions}
            \State ${q}_{b} \gets \text{Sharpen}(\bar{q}_{b},T)$ \Comment{apply temperature sharpening to the guessed label}

            \Statex
            
            \State ${h}_{x_{b,1}} \gets \mathrm{f_1}({x}^s_{b,1};\theta)$ \Comment{extract embedding of   $x^s_b$}
            \State ${h}_{x_{b,2}} \gets \mathrm{f_1}({x}^s_{b,2};\theta)$ \Comment{extract embedding of   $x^s_b$}
            \State ${h}_{u_{b,1}} \gets \mathrm{f_1}({u}^s_{b,1};\theta)$ \Comment{extract embedding of   $u^s_b$}
            \State ${h}_{u_{b,2}} \gets \mathrm{f_1}({u}^s_{b,2};\theta)$ \Comment{extract embedding of   $u^s_b$}
        \EndFor

        \Statex
        
        \State ${\mathcal{X}} \gets \{({x}^s_{b,1},{y}_b, w_b), ({x}^s_{b,2},{y}_b, w_b);b\in(1,...,B)\}$ \Comment{augmented labeled mini-batch}
        \State ${\mathcal{U}} \gets \{({u}^s_{b,1},q_{b}, w_b), ({u}^s_{b,2},q_{b}, w_b);b\in(1,...,B)\}$ \Comment{augmented unlabeled mini-batch}
        \State $\mathcal{X}^\prime, \mathcal{U}^\prime \gets \mathopen{} \text{Mixup}({\mathcal{X}}, \mathcal{U}, \alpha)$  \Comment{Apply mixup on input data}

        \Statex
        
        \State ${\mathcal{H}_x} \gets \{({h}_{x_{b,1}},{y}_b), ({h}_{x_{b,2}},{y}_b);b\in(1,...,B)\}$ \Comment{augmented labeled mini-batch}
        \State ${\mathcal{H}_u} \gets \{({h}_{u_{b,1}},q_{b}), ({h}_{u_{b,2}},q_{b});b\in(1,...,B)\}$ \Comment{augmented unlabeled mini-batch}
        \State $\mathcal{H}_x^\prime, \mathcal{H}_u^\prime \gets \mathopen{}\text{Mixup}({\mathcal{H}_x}, \mathcal{H}_u, \alpha)$ \Comment{Apply mixup on embedding space of input data}

        \Statex

        \State \Return $\mathcal{X}^\prime, \mathcal{U}^\prime, \mathcal{H}_x^\prime, \mathcal{H}_u^\prime$
    \end{algorithmic}
\end{algorithm*} 
\subsection{Semi-supervised Learning}
The presence of some noisy ID and OOD data in the fraction of clean samples results in noisy Semi-SL training. To address this issue, we propose to incorporate three components in our Semi-SL training pipeline: 1) we use the symmetric cross entropy as a robust-to-noise loss function for the supervised loss to reduce the risk of noisy label memorization; 2) we include contrastive learning to enable feature learning without labels/pseudo-labels; and 3) we use the manifold mixup as the augmentation of the feature space to enable feature learning for iterative OOD detection. This unsupervised feature learning approach significantly reduces the risk of noisy label memorization since it does not rely on the incorrect split of clean and noisy samples or on the incorrect pseudo-labels generated during Semi-SL training. Figure~\ref{fig:overall}-(c) depicts the specifics of our SSL model with semi-supervised learning on input and embedding spaces. We employ Semi-SL  by combining input mixup augmentation as well as manifold augmentation on top of the basic ideas of FixMatch algorithm, which we call ``$\mixematch$".

To achieve this, we create four copies of each sample, two with weak and two with strong random augmentations. We use the weakly augmented samples of labeled data, denoted as $x$, and unlabeled data, denoted as $u$, to create the smoothing labels and estimate the pseudo-labels for the labeled and unlabeled sets, respectively. We use the weakly augmented images to create pseudo-labels for the strongly augmented images and train the model in a semi-supervised manner on the strongly augmented images. We incorporate mixup into our pipeline similar to the MixMatch algorithm with two differences: 1) we mix strongly augmented labeled and unlabeled images with their labels, which are smoothed and guessed based on the weakly augmented labeled and unlabeled samples, respectively; and 2) we apply mixup to both the input and embedding spaces. As a result, the outputs of our $\mixematch$ algorithm are:
\begin{equation}
 \label{eq:alg}
    \mathcal{X}^\prime, \mathcal{U}^\prime, \mathcal{H}_x^\prime, \mathcal{H}_u^\prime = \mixematch(\mathcal{X}, \mathcal{U}, \mathcal{H}_x, \mathcal{H}_u, T, \alpha)\\ 
\end{equation}
where $\mathcal{X}, \mathcal{H}_x$ are input and embedding spaces of strongly augmented labeled images and  $\mathcal{U}, \mathcal{H}_u,$ are the ones for unlabeled images. Then $\mathcal{X}^\prime, \mathcal{U}^\prime$ are the mixed up augmented input space of labeled and unlabeled sets while $\mathcal{H}_x, \mathcal{H}_u$ are the mixed up augmented embedding space of labeled and unlabeled sets, respectively. The full $\mixematch$ algorithm is provided in Algorithm \ref{alg:mixematch}. In the following subsections, these augmentations and objective functions are explained.

\textbf{Mixup Augmentations:}
Applying mixup on the embedding space allows us to use semantic interpolations of the deepest layer as an additional training signal, which encourages our classifier to generate less confident predictions for interpolations of representations. Although high-level representations are often low-dimensional and useful for linear classifiers, linear interpolations of hidden representations should effectively explore significant sections of the feature space. To employ combinations of hidden data representations as a new training signal, we perform the same linear interpolation on the corresponding pair of one-hot labels, resulting in mixed samples with soft targets.

Let $x^s_1, x^s_2$ be strongly augmented images of samples $x_1, x_2$  with their corresponding pseudo-labels probabilities $p_1,  p_2$, and $h^s_1, h^s_2$ be the embedding spaces of samples $x^s_1$ and $x^s_2$. The mixed input-label pair $(x^\prime, p^\prime)$ and mixed embedding-label pair $(h^\prime, p^\prime)$ are computed as follows:
\begin{align}
   \lambda &\sim \betadist(\alpha, \alpha) \\
   \lambda^\prime &= \max(\lambda, 1 - \lambda);\label{eqn:lambda_prime} \\
   x^\prime &= \lambda^\prime x_1^s + (1 - \lambda^\prime)x_2^s; \\
   h^\prime &= \lambda^\prime h_1^s + (1 - \lambda^\prime)h_2^s; \\
   p^\prime &= \lambda^\prime p_1 + (1 - \lambda^\prime)p_2 ,
\end{align}
where $\alpha$ is a hyperparameter. Consequently, $\mathcal{X}$, $\mathcal{U}$, $\mathcal{H}_x$, $\mathcal{H}_u$ are transformed into $\mathcal{X}^\prime$, $\mathcal{U}^\prime$ and $\mathcal{H}_x^\prime$,  $\mathcal{H}_u^\prime$, which are collections of multiple mixed-up augmentations of each example with the corresponding labels.

\textbf{Loss Functions:}
The Semi-SL loss function is computed based on the strongly augmented copies as well as the mixUp augmentations of the input and embedding spaces of samples as follows:
\begin{equation}
\label{eq:loss}
\mathcal{L}^{semi} = \mathcal{L}^{sup} + \lambda_u \mathcal{L}^{unsup} + \lambda_c \mathcal{L}^{self} ,
\end{equation}
The supervised loss function ($\mathcal{L}^{sup}$) consists of two symmetric cross-entropy loss terms on the mixed-up augmented input ($\mathcal{X}^\prime$) and the mixed-up augmented embedding spaces ($\mathcal{H}_x^\prime$) of the labeled set. The unsupervised loss function ($\mathcal{L}^{unsup}$) consists of two Mean Squared loss terms on the mixed-up augmented input ($\mathcal{U}^\prime$) and the mixed-up augmented embedding spaces ($\mathcal{H}_u^\prime$) of the unlabeled set. The contrastive loss function ($\mathcal{L}^{self}$), defined in Equation (\ref{eq:loss_self}), is applied to the projection head of the embedding spaces of $\mathcal{H}_x$ and $\mathcal{H}_u$.
\section{Experiments and Results}

We evaluate our model in terms of classification accuracy on a variety of standard image classification benchmarks, such as CIFAR-10, CIFAR-100 ~\cite{2009_CIFAR}, miniImageNet~\cite{2016_NIPS_MiniImageNet}, 
Webvision~\cite{2017_arXiv_WebVision}, and Clothing1M~\cite{2015_CVPR_Clothing1M}.  


\subsection{Datasets}
\textbf{CIFAR-10/100}: The dataset used for our experiments consists of 50K and 10K training and test images with 10 and 100 image classes, respectively. Synthetic ID noise (symmetric and asymmetric) is often used to assess noise-robust algorithms on these two datasets. Our first set of experiments was conducted on a corrupted CIFAR-100, where controlled ID and OOD noise was added. We followed the same configuration proposed in~\cite{albert2022embedding}. ID noise was introduced by randomly switching the labels of $r_{in}\%$ of the training dataset to a random one. Additionally, $r_{out}\%$ of the training images were replaced with images from another dataset, either ImageNet32~\cite{2017_arXiv_IN32} or Places365~\cite{2017_TPAMI_places}, as OOD samples. By adding noise, the dataset still had the same number of images as the original CIFAR-100 dataset. 

{
\color{\COLOR}
\textbf{Mini-ImageNet}:
It consists of 50k and 10K training and test images with 100 classes, respectively. To study real-world web label noise in a controlled setting, we use Web-corrupted Mini-ImageNet from the Controlled Noisy Web Labels (CNWL) dataset~\cite{2020_ICML_MentorMix} with different noise levels (20\%, 40\%, 60\%, 80\%). 
We train our model on the red Mini-ImageNet images which are resized to $32\times32$ pixels, similar to the recent works~\cite{2021_arXiv_propmix,2021_CVPR_FaMUS,2021_arXiv_Scanmix}.
}

\textbf{(mini)Webvision:} It is constructed by using first $50$ classes of a real-world noisy Webvision dataset~\cite{2017_arXiv_WebVision}, consisting of 66k and 2.5k training and test images, respectively. We train our model on a $227\times227$ image. We report the accuracy on the mini-WebVision validation.

{
\color{\COLOR}
\textbf{Clothing1M}: It consists of one million training images gathered from online shopping websites. These images are classified into 14 different classes. There are three additional sets of clean images: 50K for training, 14K for validation, and 10K for testing. The clean training and validation sets are not utilized in our experiments, but the clean test set is used for evaluation purposes.
}
\begin{table*}[t]
    \caption{Comparison of classification accuracy with the state-of-the-art methods on ID noise and OOD noise on CIFAR-100 corrupted with ImageNet32 or Places365 images. Results of other previous methods are from~\cite{albert2022embedding}. We report best and last accuracy.
    \label{tab:cifar100}}
    \centering
    \resizebox{\textwidth}{!}
    {%
    \centering
    \begin{tabular}{c>{\centering}c>{\centering}c>{\centering}c>{\centering}c>{\centering}c>{\centering}c>{\centering}c>{\centering}c>{\centering}c>{\centering}c>{\centering}c>{\centering}c}\\
    \hline
    Corruption &$ r_{out} (\%)$& $r_{in} (\%) $& CE & Mixup~\cite{2018_ICLR_mixup}  & JoSRC~\cite{2021_CVPR_JoSRC} & ELR\cite{2020_NeurIPS_EarlyReg} & EDM~\cite{2020_WACV_EDM} & DSOS~\cite{2022_WACV_DSOS} & RRL~\cite{2021_ICCV_RRL} &SNCP~\cite{albert2022embedding} &{\bf{Ours}} \tabularnewline
    \hline
    \multirow{4}{*}{ImageNet32}& 20 & 20 & 63.7/55.5 & 66.7/62.5 & 67.4/64.2 & 68.7/68.5 & 71.0/70.4 & 70.5/70.5 & 72.6/72.3& 73.0/72.7&  \textbf{75.4/75.3} \tabularnewline
    & 40 & 20 & 58.9/44.3 & 59.5/53.2  & 61.7/61.4 & 63.2/63.1 & 61.9/61.8 & 62.5/62.1 & 66.0/65.4 &67.6/67.1& \textbf{69.1/69.1} \tabularnewline
    & 60 & 20 & 46.0/26.0 & 42.9/40.4  & 38.0/37.1 & 44.8/44.6 & 21.9/14.6 & 50.0/49.1 & 26.8/24.5 &53.4/51.3 &\textbf{59.8/59.8}  \tabularnewline
    & 40 & 40 & 41.4/18.5 & 38.4/33.9 &  41.5/41.4 & 34.8/34.2 & 24.1/01.6 & 43.7/42.9 & 31.3/30.6 &54.0/52.7& \textbf{63.0/63.0} \tabularnewline
    \hline
    \multirow{4}{*}{Places365}& 20 & 20 & 59.9/53.6 & 66.3/59.7 &  67.1/66.7 & 68.6/68.5 & 70.5/70.3 & 69.7/69.1 & 72.6/72.5  &71.3/71.1 &\textbf{74.6/74.5}\tabularnewline
    & 40 & 20 & 53.5/42.5 & 59.8/48.6  & 60.8/60.6 & 62.7/62.3 & 61.8/61.6 & 59.5/59.5  & 65.8/65.8 & 64.0/63.5 & \textbf{69.4/69.3}\tabularnewline
    & 60 & 20 & 39.6/21.4 &  39.2/33.7 &  39.8/39.6 & 37.1/36.5 & 23.7/14.7 & 35.5/35.4 &  49.3/49.3 & 49.8/49.8 & \textbf{59.0/59.0}\tabularnewline
    & 40 & 40 & 32.1/13.9 & 34.4/27.6  & 33.2/32.6 & 34.7/33.9 & 20.3/11.9 & 29.5/29.5 & 26.7/24.3 & 51.0/47.6 & \textbf{62.6/62.5}\tabularnewline
       \hline
    \end{tabular}
    }
\end{table*}

\subsection{Training Details}
We leverage PreActResNet-18 for the CIFAR-10, CIFAR-100 and mini-Imagenet datasets and PreActResNet-50 for the WebVision and Clothing1M dataset as an encoder model and a multi-layer perceptron with a single hidden layer as a projection head to project the representation onto a 128-dimensional latent space. The normalized activations of the final pooling layer $512$ and $2048$ are used as the representation vector in PreActResNet-18 and PreActResNet-50, respectively. The self-supervised model is trained for $1000$ epochs with a batch size of $1024$ on the corrupted datasets. 
Supervised training starts by filtering $10\%$ of samples detected as OOD. Then the Semi-SL step is done for $300$ epochs, where the model is trained in a supervised manner for $20$ epochs as a warmup, and training is continued in a semi-supervised way. The learning rate of the linear classifier is set at $0.1$, while the backbone is tuned with a learning rate of $0.001$ and the help of a cosine annealing scheduler. We use SGD with momentum of $0.9$ and weight decay of $5\times10^{-4}$. For OOD sample selection with KNN and ID sample selection with GMM, we set $k=100$ and $\tau_2=0.3$ for all of our experiments, respectively.
 
 \textbf{Augmentations}: We use AutoRandAugment~\cite{cubuk2019autoaugment} for SSL training. For the weak augmentation of Semi-SL step, we sequentially apply simple augmentations by random cropping followed by resizing back to the original size and random horizontal flip. For the strong augmentation, we follow the Auto-Augment policy explained in~\cite{cubuk2019autoaugment} based on CIFAR10 and ImageNet policy. The same CIFAR10-policy has been applied to both the CIFAR10/100 datasets, while the ImageNet-policy has been used for the (mini)Webvision dataset. More details are delineated in the Supplementary section.

\subsection{Results}
The average test accuracies for the CIFAR100 dataset, corrupted with ID symmetric label noise and OOD images from the ImageNet32 and Places365 datasets, are reported in Table~\ref{tab:cifar100}. Similar to~\cite{albert2022embedding,2022_WACV_DSOS}, the focus of our experiments is on OOD labeled noise. Therefore, we investigate the effect of three different levels of OOD noise rates with $r_{out} \in [20\%, 40\%, 60\%]$ and two different ratios of ID noise $r_{in} \in [20\%, 40\%]$. We show the benefits of using Manifold DivideMix to detect OOD noise data and correct the labels of ID noise samples in Table~\ref{tab:cifar100}. 

It is challenging to achieve reasonable performance at a high level of noise ratio by using a fully supervised learning model, as there are only a small number of available samples per class. However, by using SSL pre-training, Manifold DivideMix consistently shows performance improvement under different noise settings. For 40\% ID and 40\% OOD noise rates, our method achieves about 10\% improvement over state-of-the-art methods. Our method shows a high level of resilience to ID and OOD noisy labels, proving to be superior to other methods.

\textcolor{\COLOR}{In the second set of experiments, we evaluate our method on a controlled real-world label noise dataset. Jiang et al.~\cite{2020_ICML_MentorMix} established the controlled noisy web labels (CNWL) dataset, where miniImageNet is corrupted with OOD images from web queries. Table~\ref{tab:mini} illustrates the performance improvements of our method compared to the state-of-the-art methods. Our method achieves approximately a $3\%$ increase in performance on noise levels of $20\%$, and more than a $1\%$ improvement for all other noise rates. However, the SNCF method~\cite{albert2022embedding}, specifically designed for both OOD and ID noise, as well as ScanMix~\cite{2021_arXiv_Scanmix} and PropMix~\cite{2021_arXiv_propmix} using self-supervised initialization, under-perform compared to our method.}

\begin{table}[t]
    \caption{\textcolor{\COLOR}{Comparison of classification accuracy with the state-of-the-art methods on Web-corrupted miniImageNet from the CNWL~\cite{2020_ICML_MentorMix} ($32\times 32$). ``$\star$" denotes algorithms using an ensemble of networks to predict. 
    }
    \label{tab:mini}}
    \centering
    \begin{tabular}{lcccc}
    \toprule
    &\multicolumn{4}{c}{Noise Ratio}\\  \cline{2-5}  
    Method & 20\% & 40\% &60\% &80\%  \\
    \hline
    CE & 47.36& 42.70& 37.30 & 29.76 \\
    Mixup~\cite{2018_ICLR_mixup} &49.10& 46.40& 40.58& 33.58\\
    $^\star$DivideMix~\cite{2020_ICLR_DivideMix} &50.96& 46.72& 43.14& 34.50\\
    MentorMix~\cite{2020_ICML_MentorMix} &51.02 &  47.14 & 43.80 &  33.46 \\
    FaMUS~\cite{2021_CVPR_FaMUS} &51.42 &48.03 &45.10 &35.50 \\
    $^\star$ScanMix~\cite{2021_arXiv_Scanmix} & 59.06 &  54.54 &  52.36 &40.00 \\
    $^\star$PropMix~\cite{2021_arXiv_propmix}  & 61.24 & 56.22 & 52.84 &43.42 \\
    SNCP~\cite{albert2022embedding}& 61.56 & 59.94 & 54.92  & 45.62\\
    Ours & \textbf{64.4} & \textbf{61.4} & \textbf{56.2} & \textbf{47.8}\\
       \bottomrule
    \end{tabular}
\end{table}

We evaluate our proposed pipeline based on the top-1 and top-5 accuracy of (mini)WebVision and Clothing1M reported in Table~\ref{tab:cloth-webvision}.
We believe that by using the self-supervised pre-training step and considering contrastive loss during the semi-supervised step, the model learns more generalizable features, which reduces the risk of overfitting to noisy samples as well as overconfident prediction on the semantically different class samples in the noisy real-world dataset. Our approach achieves superior results compared to the recent GSS-SSL method~\cite{yu2023prevent}, establishing a new state-of-the-art top-1 accuracy for (mini)WebVision. However, it is worth noting that the GSS-SSL method outperforms ours on the Clothing 1M dataset.

\begin{table}[]
\caption{\textcolor{\COLOR}{Comparison of classification accuracy with the state-of-the-art methods on Clothing1M and (mini)Webvision using ResNet50 as the backbone model.}
\label{tab:cloth-webvision}}
\begin{tabular}{lccc}
\hline
\multicolumn{1}{l}{\multirow{2}{*}{Method}} & \multicolumn{1}{l}{Clothing1M} & \multicolumn{2}{l}{(mini)WebVision} \\  \cline{2-4} 
\multicolumn{1}{c}{}   & Top1  & Top1 & Top5          \\ \hline
GCE~\cite{2018_NeurIPS_GCE}  & 71.7   & 61.2& 80.8          \\
SL~\cite{2019_ICCV_SL}  & 72.1   & 63.8   & 84.3          \\
ELR+~\cite{2020_NeurIPS_EarlyReg} & 71.5   & 63.6   & 83.5          \\
Co-teaching~\cite{2018_NeurIPS_CoTeaching}   & 72.5   & 64.1  & 85.0         \\
JoCor~\cite{2021_CVPR_JoSRC}  & 71.7  & 60.8 & 82.5          \\
DivideMix~\cite{2020_ICLR_DivideMix}    & 74.6    & 77.2  & 91.6          \\
GSS-SSL~\cite{yu2023prevent}   & \textbf{74.9}  & 77.4    & \textbf{93.1} \\
Ours    &      73.1       & \textbf{78.4} & 92.0         \\
\hline
\end{tabular}
\end{table}

\subsection{Ablation Studies}

First, we analyze our method under the ID label noise scenario applied in CIFAR10 dataset under different symmetric noise rates of 20\%, 50\%, 80\%, and 90\%, as well as asymmetric noise rates of 10\%, 30\%, and 40\% in Table~\ref{tab:cifar10}. Our method outperforms the baseline methods for the cases of 50\% and 80\% symmetric noise rates, and achieves a significantly better performance over the state-of-the-art for 90\% noise rate. For high noise rates, loss-based selection procedures \cite{2020_ICLR_DivideMix} often fail due to the selection of a large number of noisy samples as clean ones. As a result, SSL pre-training helps our method, as well as PropMix~\cite{2021_arXiv_propmix}, to not struggle in teh presence of sever label noise.

Furthermore, we investigate the asymmetric noise scenario, where each class is not equally affected by label noise, leading to a more complicated selection process. Therefore, at high levels of label noise, the balancing approach in UNICON and PropMix achieves better performance. However, for low noise rates, our method performs slightly better than the others. 

\begin{table}[t]
\caption{Comparison of classification accuracy with the state-of-the-art methods on CIFAR10 with ID symmetric and asymmetric noise.}
 \label{tab:cifar10}
\centering
\scalebox{0.85}{
\begin{tabular}{lccccccc} 
\toprule
&  \multicolumn{4}{c}{\textbf{ Symmetric Noise}}  &  \multicolumn{3}{c}{\textbf{ Asymmetric Noise}}  \\  \cline{2-8}  
Method  &  20\%  &  50\%  &  80\%  &  90\%  &  10\%  &  30\%  &  40\%  \\  \hline
  
{CE}   &  86.8  &  79.4  &  62.9  &  42.7  &  88.8  &  81.7  &  76.1 \\ 
{MixUp \cite{2018_ICLR_mixup}}   &  95.6  &  87.1  &  71.6  &  52.2 &  93.3  &  83.3  &  77.7  \\
{JPL \cite{kim2021joint}} &  93.5  &  90.2  &  35.7  &  23.4 &  94.2  &  92.5  &  90.7  \\
{MOIT \cite{2021_CVPR_MOIT}}  &  94.1  &  91.1  &  75.8  &  70.1 &   94.2  &  94.1  &  93.2 \\ 
{DivideMix \cite{2020_ICLR_DivideMix}}   &  95.0 &  93.7  &  92.4  &  74.2   &  93.8  &  92.5  &  91.4  \\
{ELR \cite{2020_NeurIPS_EarlyReg}}  &  95.8  &  94.8  &  93.3  &  78.7  &   95.5  &  94.8  &  93.0 \\ 
{UNICON \cite{karim2022unicon}}  &    96.0  &  95.6  &  93.9  &  88.1  &  95.3  &  94.8  &  94.1  \\ 
{Propmix \cite{2021_arXiv_propmix}}  &    \textbf{96.1}  &  95.5  &  93.7  &  93.2  &  -  &  -  &  \textbf{94.6}  \\ 
{GSS-SSL \cite{yu2023prevent}}   &  94.3 &  -  &  91.6  &  -   &  -  &  92.4  &  91.8 \\
\textbf{Ours} &   96.0  &  \textbf{95.7}  &  \textbf{94.6}  &  \textbf{93.7}   &   \textbf{95.6}  &  \textbf{95.0}  &  93.8  \\  
\bottomrule

\end{tabular}
}
\end{table}


In Table~\ref{tab:abla}, we evaluate the impact of each component of our proposed method by investigating scenarios where we only apply sample selection for ID noise, as well as the combined impact of the ID and OOD sample selection (denoted as ID + OOD). In this experiment, we demonstrate the effectiveness of our MixEMatch algorithm by comparing it to the FixMatch algorithm in terms of model generalization during test time, specifically by incorporating both input and embedding mixup techniques. Unlike FixMatch, MixEMatch applies mixup to both input and embedding spaces, whereas FixMatch only applies mixup on the input space. In the Semi-SL step, the MixEMatch algorithm yields considerable gains (about $5\%$ accuracy) for the Manifold DivideMix model. This implies that MixEMatch takes advantage of the Manifold mixup idea, which is useful for both ID and OOD sample selection algorithms.

\begin{table}[t]
    \caption{Ablation study on CIFAR-100 corrupted with ImageNet32 with $r_{out} = 40\%$ and $r_{in} = 40\%$. \label{tab:abla}}
    \global\long\def\arraystretch{1}%
    \centering
    \resizebox{0.45\textwidth}{!}{{{}}%
    \begin{tabular}{l l>{\centering}c>{\centering}c>{\centering}c>{\centering}c}
    \toprule
        & & FixMatch & MixEMatch  & Best & Last\tabularnewline
        \hline
       \multirow{2}{*}{No noise corr} & CE & \xmark & \xmark & 41.4  &18.5 \tabularnewline
       & SSL+LC & \xmark & \xmark & 47.5 &19.9\tabularnewline
       \hline
       \multirow{5}{*}{Noise Robust} 
       & ID  &  \cmark & \xmark & 58.6  & 58.3\tabularnewline
       & ID + OOD &   \cmark & \xmark &   58.0& 58.0 \tabularnewline
       \cmidrule(lr){2-6}
       & ID &  \xmark & \cmark & 61.6 & 61.4\tabularnewline
       & ID + OOD &   \xmark & \cmark & \textbf{63.1} & \textbf{63.0}  \tabularnewline
       \hline
    \end{tabular}}

\end{table}

\section{Conclusion}
We have developed a semi-supervised framework that leverages self-supervised learning to address both in- and out-of-distribution label noise. To achieve this, we first train a deep neural network model using unsupervised contrastive learning on a noisy dataset to extract robust and flexible embedding representations of the input data that are not limited by explicit labels. Then, we use a basic K-Nearest Neighbors algorithm in the embedding space to identify likely out-of-distribution samples and exclude them from the subsequent semi-supervised learning stage. We also add a linear classifier on top of the self-supervised model to identify in- and out-of-distribution label noise, as well as clean sample sets, based on the clustering of loss values. To further improve the generalization performance and representation learning, we propose MixEMatch, an algorithm that applies mixup augmentation in both the input and representation spaces during the semi-supervised learning step. These augmentations enhance the overall quality of pseudo-labels and significantly improve the effectiveness of semi-supervised learning.
Our experiments show that unsupervised feature learning reduces the effects of overfitting to label noise and improves noisy sample selection, particularly in the presence of severe noise. Our approach outperforms recent methods for learning with noisy labels and offers more consistency across varying noise levels. For example, with 40\% in- and 40\% out-of-distribution label noise, our model improves the accuracy of corrupted CIFAR-100 over $10\%$.  Additionally, our proposed approach achieves state-of-the-art top1 accuracy on the WebVision dataset.  

{
\small
\bibliographystyle{ieee_fullname}
\bibliography{references}
}

\end{document}